%% file: paper.tex
\documentclass{article}
\PassOptionsToPackage{numbers,sort&compress}{natbib}

\usepackage[preprint]{neurips_2026}

\usepackage[utf8]{inputenc} 
\usepackage[T1]{fontenc}    
\usepackage{hyperref}       
\usepackage{url}            
\usepackage{booktabs}       
\usepackage{amsfonts}       
\usepackage{nicefrac}       
\usepackage{microtype}      
\usepackage{xcolor}         

\usepackage{hyperref}
\usepackage{algorithm}
\usepackage{algpseudocode}

\usepackage{amsmath}
\usepackage{amssymb}

\usepackage{xcolor, soul}
\sethlcolor{pink}

\usepackage{amsmath}
\usepackage{amssymb}
\usepackage{mathtools}
\usepackage{amsthm}

\usepackage{booktabs}
\usepackage{booktabs}
\usepackage{multirow}
\usepackage{adjustbox}
\usepackage{xcolor}
\usepackage{colortbl}
\usepackage[table]{xcolor}
\definecolor{lightgray}{gray}{0.9}
\usepackage{pifont} 
\usepackage{multirow}
\usepackage{wrapfig}

\newcommand{\cmark}{\ding{51}} 
\newcommand{\xmark}{\ding{55}} 
\usepackage{tabularx}

\definecolor{Gray}{gray}{0.9}

\newcommand{\eg}{\textit{e.g.}}
\newcommand{\ie}{\textit{i.e.}}

\definecolor{tsneblue}{RGB}{80,136,183}
\definecolor{tsnepurple}{RGB}{189,131,180}
\usepackage[capitalize,noabbrev]{cleveref}

\usepackage[table]{xcolor}
\usepackage{subcaption}

\usepackage{pgfplots}
    \usepgfplotslibrary{colorbrewer}
    \pgfplotsset{
        cycle list/Dark2,
        cycle multiindex* list={
            mark list*\nextlist
            Dark2\nextlist
        },
    }
\pgfplotsset{compat=1.14}
\definecolor{Cerulean}{rgb}{0.0, 0.48, 0.65}
\definecolor{goldenpoppy3}{rgb}{0.99,  0.46, 0.1}
\definecolor{ForestGreen}{RGB}{34,139,34}
\definecolor{darkpastelpurple2}{rgb}{0.68, 0.33, 0.83}
\definecolor{caribbeangreen2}{rgb}{0.0, 0.825, 0.625}
\definecolor{OrangeRedSoft}{RGB}{240,40,50}
\definecolor{OliveDrabSoft}{rgb}{0.55, 0.62, 0.20}   
\definecolor{DeepRose}{rgb}{0.78, 0.18, 0.45}       
\definecolor{darkcerulean2}{rgb}{0.00, 0.20, 0.35}

\theoremstyle{plain}

\theoremstyle{definition}

\theoremstyle{remark}

\title{Extending Pretrained 10-Second ECG \\Foundation Models to Longer Horizons}

\author{%
{\bfseries
Wei Tang$^{1}$ \enspace 
Jinpei Han$^{2}$ \enspace
Kangning Cui$^{3}$ \enspace
Mattia Carletti$^{6}$ \enspace
Fredrik K. Gustafsson$^{6}$} \\
{\bfseries
Shreyank N Gowda$^{4}$ \enspace
Patitapaban Palo$^{6}$ \enspace
Anshul Thakur$^{6}$ \enspace
Lei Clifton$^{6}$} \\
{\bfseries
Jean-michel Morel$^{5}$ \enspace
Raymond H. Chan$^{5}$ \enspace
David A. Clifton$^{6}$ \enspace
Xiao Gu$^{6}$} \\
\\[-0.5em]
$^{1}$City University of Hong Kong \enspace
$^{2}$Imperial College London \enspace
$^{3}$Wake Forest University \\
$^{4}$University of Nottingham \enspace
$^{5}$Lingnan University \enspace
$^{6}$University of Oxford
}

\begin{document}

\maketitle

\begin{abstract}
    Electrocardiogram (ECG) foundation models pretrained on typical diagnostic 10-second ECG segments, have demonstrated strong transferability across a range of clinical applications. However, many real-world applications produce recordings that are typically longer, and are varied in duration during inference time. These 10-second models have no built-in way to combine information across time. Extending them to longer horizons introduces two challenges: \textit{structural} incompatibilities arising from input-length disparities, and \textit{semantic} challenges that limit meaningful temporal aggregation. We propose a parameter-efficient framework that extends pretrained ECG foundation models to longer and variable-length ECGs without retraining the backbone. Guided by a frozen pretrained 10-second model, we introduce a lightweight plug-in module that extends the model in two complementary ways: (i) structurally compatible long-sequence processing and (ii) semantically informed temporal modeling. Experiments on multiple long-horizon ECG tasks, datasets, and foundation model backbones demonstrate that our method enables robust long-horizon extension from pretrained snapshot models, consistently outperforming sliding-window and pooling-based baselines with strong parameter efficiency.
\end{abstract}

\section{Introduction}
\input{section/introduction}

\vspace{-5pt}

\section{Related Work}
\input{section/related_work}

\vspace{-5pt}

\section{Method}

\input{section/method}

\vspace{-5pt}

\section{Experiments}
\label{experiments}

\input{section/experiment}

\section{Results}
\input{section/results}

\vspace{-8pt}
\section{Conclusion}
\input{section/conclusion}

\bibliographystyle{unsrtnat}  
\bibliography{reference} 

\clearpage


\end{document}

%% file: section/introduction.tex

The electrocardiogram (ECG) remains the gold standard for non-invasive cardiovascular diagnosis, providing a direct view of cardiac electrical activity. Advances in machine learning, particularly deep neural networks, have substantially improved automated ECG analysis, with models achieving expert-level performance on several diagnostic tasks \cite{siontis2021artificial}. Early deep learning approaches, however, primarily relied on supervised training for narrow, task-specific objectives. While effective in controlled settings, these models were constrained by limited labeled data and often failed to generalize across datasets, institutions, and patient populations \cite{strodthoff2020deep}.

To improve robustness and transferability, recent work has increasingly shifted toward ECG foundation models pretrained with self-supervised objectives on large-scale ECG datasets \cite{li2024electrocardiogram, gu2025bfm, mckeen2025ecg}. Through contrastive or generative self-supervised learning objectives, these architectures learn generic feature extractors that are robust to signal variability and improve transfer across cohorts and institutions. As a result, a single pretrained foundation model backbone paired with lightweight task-specific heads can be adapted to diverse downstream applications such as diagnosis, phenotyping, and outcome prediction \cite{gu2025sensing,li2025electrocardiogram}. 


Despite their progress, these ECG foundation models present a fundamental limitation: they are trained and used as fixed-length \textit{snapshot} processors. They typically operate on short inputs (often 10\,s) during both pretraining and inference \cite{li2024electrocardiogram, gu2025bfm}, which biases the learned representations toward local waveform morphology and short-term rhythm. As a result, they lack an explicit mechanism for aggregating information across extended recordings, modeling long-range temporal dependencies, or supporting coherent prediction across multiple horizons -- when recording duration varies across monitoring scenarios. 

\begin{figure}[tp]
    \vspace{-5mm}
    \centering
    \includegraphics[width=\textwidth]{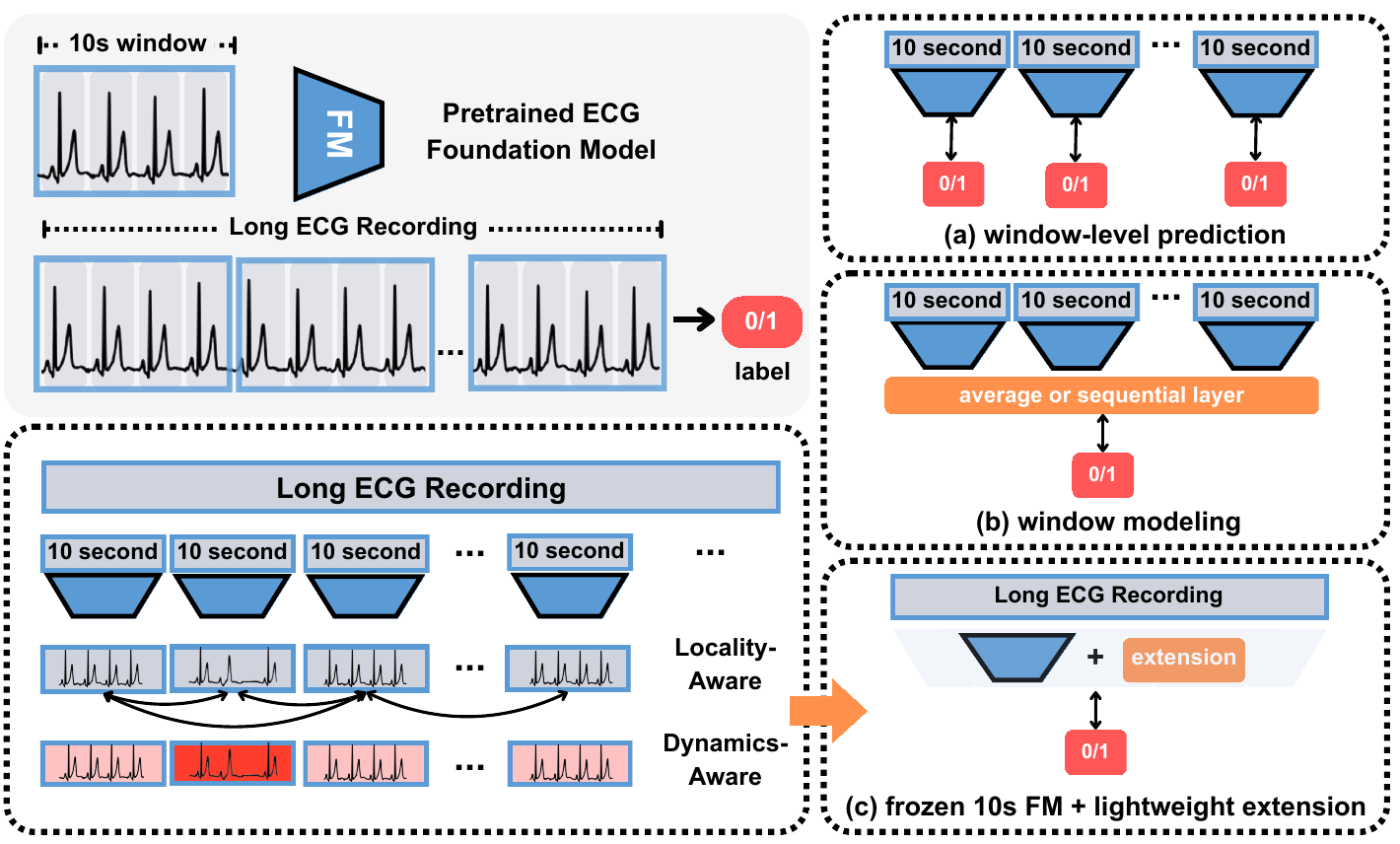}
    \caption{\textbf{Overview of the problem setting and extension strategies.}
ECG foundation models are typically pretrained on short, fixed-length recordings (\eg{}, 10\,s), which makes direct use on long or variable-length recordings non-trivial.
Naive extension strategies either run independent window-level predictions and aggregate the outputs as panel (a), or apply simple aggregation or sequential layers over window representations as panel (b).
We propose a lightweight long-horizon extension for frozen 10-second ECG foundation models that supports variable-length longer recordings, by leveraging locality-aware inter-window similarity and dynamics-aware long-term learning, as illustrated in panel (c).}
\vspace{-10mm}
    \label{fig:overview}
\end{figure}

This limitation is consequential for many clinically relevant settings. In contrast to short 12-lead diagnostic recordings, continuous monitoring is intended to capture sparse or intermittent events over longer horizons, often under reduced-lead configurations \cite{lehman2023vtac,PhysioNet-mimic3wdb-matched-1.0,kansal2025mcmed}. In these contexts, clinically actionable signals may be distributed unevenly over time, as in paroxysmal atrial fibrillation surveillance, sleep, or sepsis early-warning. A common workaround is sliding-window inference followed by simple aggregation (\eg{}, mean pooling) of window-level outputs (Figure~\ref{fig:overview}a). More structured variants aggregate window representations using attention or lightweight sequential modules (Figure~\ref{fig:overview}b). While practical, such strategies can be computationally costly and still model cross-window dependencies only indirectly, motivating adaptation methods that explicitly extend the model temporal horizon beyond the original snapshot setting. 

In this work, we show that pretrained snapshot ECG foundation models can be adapted to longer-duration and variable-length inputs without computationally expensive retraining. Our key observation is that these models can be repurposed for extended-context ECG analysis along three complementary dimensions, as illustrated in \figureautorefname{}~\ref{fig:overview} (left bottom panel).

\noindent\textbf{General ECG interpretation.}
Pretrained ECG foundation models already acquire substantial domain knowledge for interpreting cardiac signals through large-scale pretraining. This enables knowledge to be preserved and effectively reused when extending the models to longer temporal horizons, without relearning low-level ECG semantics.

\textbf{Locality-aware representation learning.}
Pretrained ECG foundation models are already well suited to encode short 10-second ECG windows: each segment can be meaningfully represented on its own, thus can be used as a local reference when extending the model to longer recordings. We exploit this property to preserve local semantics and align snapshot-level representations when processing long recordings.

\noindent\textbf{Dynamics-aware long-term learning.}
The representations produced by pretrained models naturally reflect relative importance across segments, offering a basis for identifying informative events within long recordings. This property allows selective reuse and robust aggregation of local representations, which is essential for modeling sparse and transient dynamics over extended time horizons.

Based on these insights, we propose a parameter-efficient framework that extends frozen snapshot ECG foundation model to longer-horizon and variable-length settings. Rather than modifying or retraining the pretrained backbone, we reuse its embedded ECG knowledge and locality prior, while introducing a lightweight module to explicitly model long-term temporal dynamics. Across diverse datasets, tasks, and models, our method consistently improves long-horizon performance over conventional sliding-window and pooling strategies, demonstrating a practical and scalable path that goes beyond 10-second ECG understanding.

%% file: section/related_work.tex
\noindent\textbf{ECG foundation models.} Foundation models have recently been extended from vision and language to physiological signals, with ECG serving as a prominent testbed. Most existing ECG foundation models are pretrained on diagnostic ECG recordings, which are typically short in duration ($\approx$10\,s) and acquired using standard 12-lead configurations \cite{li2024electrocardiogram, gu2025bfm}.
To date, most progress in this field has been driven along two complementary dimensions: scaling the volume and diversity of pretraining data \citep{wan2025openecg,li2025electrocardiogram,moody2025foundation,gu2025sensing}, and refining pretraining strategies to better exploit available supervision (self-supervised \cite{song2025crema}, hybrid \cite{zhang2025ecgfm}, or knowledge-informed objectives \cite{yu2024ecg,tian2024foundation}). On the other hand, these models, pretrained in a diagnostic 10-second 12-lead regime, have been shown to generalize seamlessly to reduced-lead or single-lead settings~\cite{gu2025sensing,li2025electrocardiogram}, which are widely used in continuous monitoring scenarios. However, they remain fundamentally limited in their ability to operate directly over longer temporal horizons, in an end-to-end manner, beyond simplistic sliding-window processing of 10\,s segments.

\noindent\textbf{Long-term physiology modeling.} Many clinically relevant decisions depend on physiological dynamics over long horizons, which are not well captured by short snapshot windows, such as ambulatory Holter monitoring and intensive care monitoring. Recent learning-based systems often apply window slicing, which partitions a long recording into short segments, produces segment-level predictions or embeddings, and then aggregates these across time. The aggregation strategy varies from simple pooling to the more recent attention pooling or a lightweight sequential layer like LSTM to model the temporal relationships \cite{zheng2021predicting, zhang2023automatic}. However, such aggregation introduces a structural bottleneck: As most tasks are formulated as mapping a single long sequence to a single outcome, temporal dependencies within the sequence are often weakly modeled, making it difficult to localize when clinically meaningful risk signals occur. Moreover, performance can degrade when predictions are made from observations that are shorter than the horizon assumed by the aggregation, limiting utility in time-sensitive clinical decision-making.

\noindent\textbf{Adaptation of biosignal foundation models.} Alongside the emergence of ECG foundation models, or broader biosignal foundation models, increasing attention has been paid to adaptation under distribution shifts \cite{wang2025neurottt} and deployment constraints \cite{zhou2025htuning}. They are mostly enabled by parameter-efficient fine-tuning methods such as LoRA, prefix tuning, and adapter modules, which update only a small subset of parameters while largely preserving the backbone \cite{wu2025efficient}. Despite this progress, existing biosignal adaptation work primarily targets task transfer, domain shift, and personalization under a fixed input scale. It assumes that the downstream signal can be processed within the same short window length used in pretraining. Due to the nature of most existing ECG datasets, which were typically collected in standard diagnostic settings (often 10\,s), the models pretrained on these datasets fall short when applied to longer temporal horizons required in several clinical settings. A systematic approach to extend the temporal horizon of pre-trained biosignal foundation models during adaptation remains unexplored.

%% file: section/method.tex
We focus on ECG foundation models based on transformer architectures and denote by $f_{\theta}$ the backbone foundation model, parameterized by $\theta$, whose parameters are kept frozen. The input to the backbone model is an ECG signal, processed as a sequence of fixed-length patch-level embeddings (\eg{}, $0.1\, s$), which are augmented with positional embeddings to encode temporal order before being passed to the transformer. Specifically, the ECG input embedding is denoted as $\mathbf{X} \in \mathbb{R}^{N_{[10s]} \times D}$, where $N_{[10s]}$ is the number of patches and $D$ is the embedding dimension. The corresponding positional embeddings $\mathbf{E}_{[10s]} \in \mathbb{R}^{N_{[10s]} \times D}$, are added to $\mathbf{X}$ and passed to the backbone model $f_{\theta}$ to produce a global representation of the signal, $f_{\theta}(\mathbf{X}+\mathbf{E}_{[10s]})$.
Existing ECG foundation models are typically pretrained on short, fixed-length segments (\eg, 10-second windows) and achieve strong performance by learning generalizable representations of local waveform morphology and short-term rhythm. However, applying these models directly to longer horizons poses two fundamental challenges.

\vspace{-10pt}
\begin{itemize}
    \item \textbf{Structural extension.} At a structural level, snapshot ECG models are not directly compatible with longer recordings. Their positional encodings, tokenization schemes, and input assumptions are optimized for short segments.
    \vspace{-5pt}
    \item \textbf{Semantic extension.} Beyond input compatibility, long-horizon ECG analysis requires meaningful aggregation of information across time. Clinically relevant events may be sparse, transient, or distributed unevenly over long recordings. Naively processing each 10-second segment independently and pooling predictions fails to capture long-range dependencies.
\end{itemize}
These two challenges motivate a framework that not only enables snapshot ECG foundation models to operate on long recordings, but also extends their semantic understanding to support coherent long-term modeling. To this end, we propose a framework, illustrated in Figure~\ref{fig:method_overview}, that performs a single adaptation to long-horizon ECG signals ($\mathbf{X} \in \mathbb{R}^{N_{[L]} \times D}$, with $L > 10\,\mathrm{s}$) and supports inference across variable signal lengths up to $L$ without further modification. The empirical value for $L$ considered in the main experiments is 180\,s.

\input{figures/method2}

\vspace{-5pt}

\subsection{Structural Extension - Enabling Long-Horizon ECG Processing} 
\label{subsec:pos_encoding}

A critical barrier to adapting snapshot models to extended recordings lies in the incompatibility of positional embeddings. These embeddings $\mathbf{E}_{\text{[10s]}} \in \mathbb{R}^{N_{[10s]} \times D}$ are learned for a fixed input length and therefore encode temporal relationships that are specific to a 10-second window. 

In the image domain, a common practice for extending vision transformers to different image scales \cite{dosovitskiy2021an, bao2022beit, liu2021Swin}, is interpolating positional embeddings. However, this strategy does not naturally transfer to time series, especially biosignals, since interpolation has fundamentally different meanings between these two domains. The former (image) is related to resolutions, whilst the latter (biosignal) is related to the sampling rate, which is significantly entangled with their physiological information.

To address this, rather than interpolating from $N_{[10s]}$ to $N_{[L]}$, we introduce a two-level positional encoding scheme, comprising local (fine-grained, the original $\mathbf{E}_{[10s]}$) and global (coarse-grained, $\mathbf{E}_{\text{global}}$) components. The former reuses the original $\mathbf{E}_{[10s]}$ to preserve within–10-second temporal structure, whereas the latter models the relationship between consecutive 10-second segments. For a sequence of $N_{[L]}$ patches, we construct the final positional embedding $\mathbf{E}_{\text{[L]}} \in \mathbb{R}^{N_{[L]} \times D}$ as:
\begin{equation}
\small
    \mathbf{E}_{\text{[L]}}[t] =  \mathbf{E}_{[10\text{s}]}\!\left[t \bmod N_{[10\text{s}]}\right] + \mathbf{E}_{\text{global}}[\lfloor t / {N_{[10s]}}\rfloor],
    \label{eq:pos_encoding}
\end{equation}
where $t$ indexes the patch and $\lfloor t / {N_{[10s]}} \rfloor$ indexes its associated 10-second segment for global temporal encoding. 

\textbf{Local component ($\mathbf{E}_{\text{[10s]}}$).} This component inherits the pre-trained positional weights, preserving sensitivity to high-frequency intra-segment structures (\eg{}, P-QRS-T complexes). For sequences exceeding the pretraining length, we repeat these weights rather than interpolating them, preserving the temporal scale at which local waveform morphology was learned during pretraining.

\textbf{Global component ($\mathbf{E}_{\text{global}}$).} To anchor local windows within the broader recording context, we introduce a learnable embedding matrix $\mathbf{E}_{\text{global}} \in \mathbb{R}^{M \times D}$, where $M=N_{[L]} / N_{[10s]}$ denotes the number of 10-second segments in the signal. For simplicity and without loss of generality, we assume that $N_{[L]}$ is an integer multiple of $N_{[10s]}$.
Each embedding in $\mathbf{E}_{\text{global}}$ is then broadcast across a contiguous block of $N_{[10s]}$ patches by \equationautorefname{}~\ref{eq:pos_encoding}. 

This hierarchical design enables modeling over the full recording duration while maintaining the frequency characteristics and morphological semantics learned within local segments during pretraining.

\textbf{Soft prompt tuning.} To facilitate the learning of global positional embeddings and enable parameter-efficient adaptation to capture long-horizon temporal relationships, we adopt soft prompt tuning by introducing additional learnable tokens that are prepended to the input sequence and inserted at intermediate layers \cite{jia2022visual}. 

\subsection{Semantic Extension - Extending Long-Horizon ECG Understanding}
\label{subsec:semantic_extension}
A purely structural extension to longer signals does not, by itself, work directly on longer signals, as the additional components introduced by the extension are randomly initialized. Directly and solely optimizing learnable components in an end-to-end manner, by mapping the long-sequence to a single label, may results in suboptimal extension and may be insufficient for effective long-range modeling. 

To address this, we take the hypothesis that the backbone foundation model $f_{\theta}$ itself already learns representations that are sufficiently informative at any 10-second scale locally, and when applied to consecutive 10-second segments, can be leveraged to capture dynamics over longer horizons. 
In this sense, we adopt a teacher-student framework, where the representation derived from each 10-second segment serves as the teacher to guide the representation learning both locally and globally over extended temporal horizons. 

\textbf{Locality-aware representation learning.} We first enforce segment-level semantic alignment between the pretrained and extended models. The pretrained foundation model is trained on short (10-second) ECG segments and provides robust local representations of morphology, rhythm, and physiological state. When adapting the model to longer ECG sequences, these local representations may be changed by the long-horizon training objective. To reduce this risk, we constrain the student model to remain aligned with the pretrained teacher at the segment level.

We implement this using teacher-guided soft contrastive learning. Prior approaches define contrastive pairs based on temporal adjacency \cite{lee2024soft} or patient identity \cite{kiyasseh2021clocs}. In contrast, we compute soft similarity targets from teacher-derived representations within each batch, thereby preserving the teacher-induced similarity structure without imposing explicit temporal or patient-level pairing assumptions.


We use pairwise similarities between teacher representations as soft supervision to guide the student via a contrastive objective. Let $\mathbf{S}^T_{i,j}$ denote the similarity (\eg{}, negative $L_2$ norm) between teacher representations of local 10-s segments $i$ and $j$. For an anchor segment $i$, if $\mathbf{S}^T_{i,j} > \mathbf{S}^T_{i,k}$, the student is encouraged to preserve the same ordering, \ie{}, $\mathbf{S}^S_{i,j} > \mathbf{S}^S_{i,k}$. This is implemented using a rank-based soft contrastive loss \cite{zha2023rank}, which preserves the relative ranking of inter-segment similarities. \figureautorefname{}~\ref{fig:method_overview} (right upper panel) illustrates this similarity-based alignment, with the local contrastive loss formulated as below:
\begin{equation}
\small
    \mathcal{L}_{\text{local}}
    = - \frac{1}{K}
    \sum_{\substack{j=1 \\ j \neq i}}^{K}
    \log
    \frac{\exp(\mathbf{S}^S_{i,j}/\tau)}
    {\displaystyle \sum_{k \in \Omega_{i,j}} \exp(\mathbf{S}^S_{i,k}/\tau)
    + \exp(\mathbf{S}^S_{i,j}/\tau)},
\label{eq:local}
\end{equation}
where $K$ denotes the number of segments in the batch, and $\tau$ is a temperature parameter. The set $\Omega_{i,j}$ is defined based on the teacher similarity as $\Omega_{i,j} = \{ k \mid \mathbf{S}^{T}_{i,k} < \mathbf{S}^{T}_{i,j} \}$,
\ie{}, all segments that are ranked as less similar to anchor $i$ than segment $j$ according
to the teacher.
In practice, we add a projector $g_\text{local}(\cdot)$ on top of the latent representation to implement this loss, following standard practice in contrastive learning \cite{chen2020simple}.

\textbf{Dynamics-aware consistency learning.}
Physiological signals such as ECG are inherently non-stationary, exhibiting pronounced temporal variability over long recordings. Rather than assuming that all 10-second segments should contribute in the same way, we use the frozen teacher representation to partition segments according to their similarity structure. This partition is used as an operational mechanism for organizing heterogeneous segments (such as rare abnormal events versus stationary patterns, or clean versus less clean segments), and empirically, we found that using both groups consistently outperforms using either group alone, which indicates that both high-variation and lower-variation parts contain useful information. We therefore encourage each group to retain information from the whole recording, to capture global dynamics.


To implement this idea, we identify a subset of 10-second segments $\mathcal{I}_{bg}$ that are most consistent with the rest of the recording, using an affinity matrix computed by the frozen teacher encoder. Intuitively, the affinity matrix encodes the pairwise similarity between all segments, and high-affinity (ha) segments $\mathcal{I}_{ha}$ are selected by ranking across the affinity matrix. We then encourage these high-affinity segments to remain aligned to the global representations, remaining as consistent as possible even when the more dynamic segments are excluded. 
Specifically, the consistency is achieved as below:
\begin{equation}
    \mathcal{L}_{\text{dyn}} = \frac{1}{|\mathcal{I}_{ha}|} \sum_{i \in \mathcal{I}_{ha}} \mathcal{D}_{\text{cos}} \left( g_\text{dyn}(\mathbf{z}_S^{(i)}), \text{stopgrad}(\bar{\mathbf{z}}_S) \right),
\label{eq:dyn}
\end{equation}
where $\mathbf{z}_S^{(i)}$ denotes the student feature for 10-second segment $i$, $\bar{\mathbf{z}}_S$ is the global feature by averaging latent features over the
entire long-horizon sequence, and $\mathcal{D}_{\text{cos}}$ is the cosine distance. In practice, we apply a one-sided $g_\text{dyn}(\cdot)$ projection head to the student representations, and a stop-gradient $\text{stopgrad}(\cdot)$ operator through global representation, to prevent trivial collapse to constant representations~\cite{chen2021exploring}.

\subsection{Model Training}
\label{sec:optim}

We train the framework end-to-end by minimizing:
\begin{equation}
    \mathcal{L} = \mathcal{L}_{\text{task}} + \lambda_1 \mathcal{L}_{\text{local}} + \lambda_2 \mathcal{L}_{\text{dyn}},
    \label{eq:full_loss}
\end{equation}
where $\mathcal{L}_{\text{task}}$ is the standard classification loss, $\lambda_1$ and $\lambda_2$ are scalar weighting coefficients that balance the contribution of each auxiliary loss. 

%% file: figures/method2.tex
\begin{figure*}[t]
    \centering
    \includegraphics[width=\linewidth]{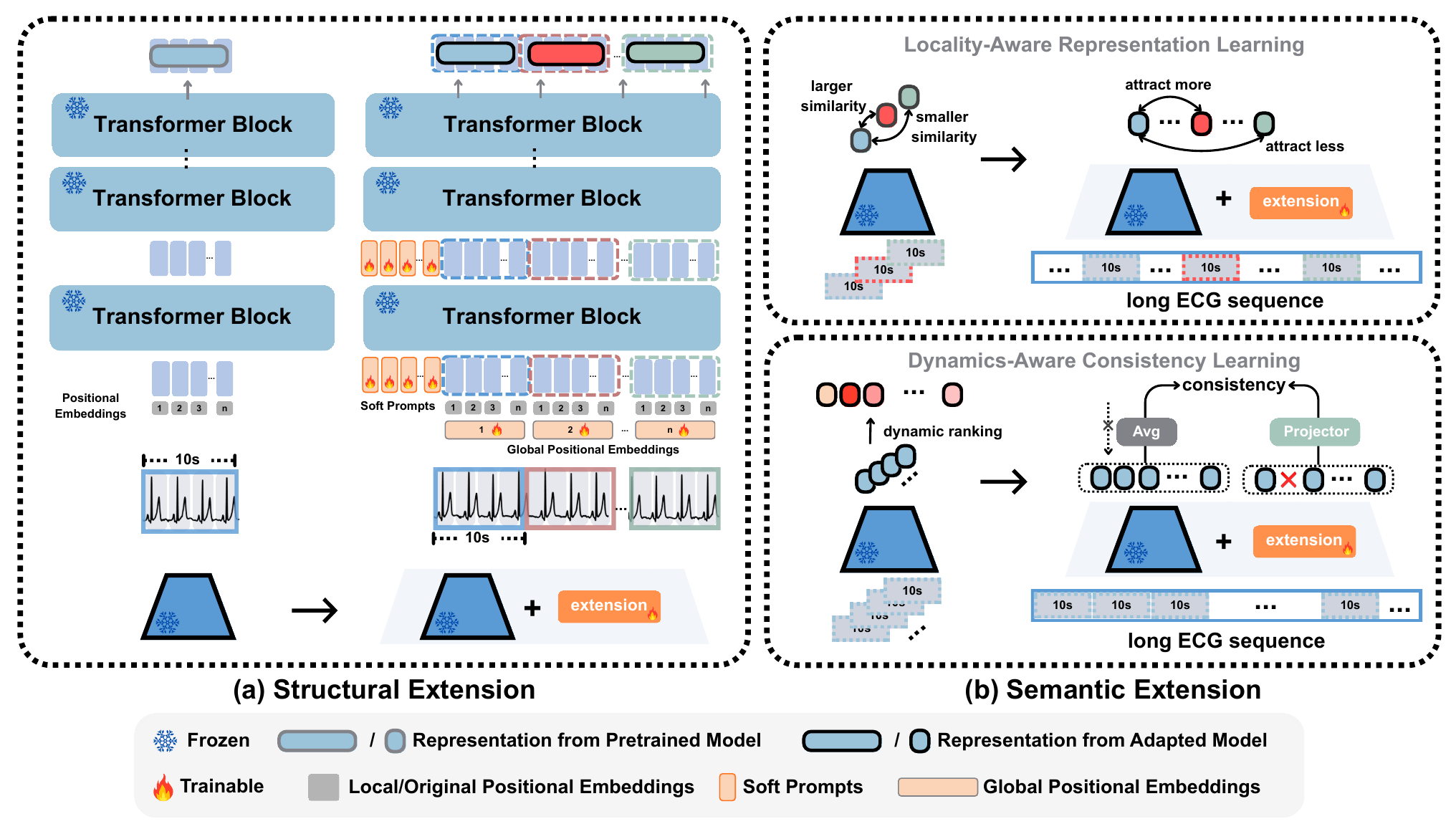}
    \caption{\textbf{Overview of our extension framework.} The extension is divided into two complementary parts: a structural extension that enables compatibility with long-horizon recordings, and a semantic extension that supports coherent representation learning over extended temporal horizons. (a) Structural extension (Section~\ref{subsec:pos_encoding}) introduces additional learnable tokens, soft prompts and global positional embeddings, by freezing the original backbone model, to make it compatible with longer recordings. (b) Semantic extension (Section~\ref{subsec:semantic_extension}) leverages snapshot-level (10-second) representations derived from the frozen backbone as supervision, enforcing semantic consistency across segments and supporting coherent long-horizon understanding.}
    \label{fig:method_overview}
    \vspace{-15pt}
\end{figure*}

%% file: section/experiment.tex
We evaluate the proposed framework on four downstream task settings and four foundation model backbones. This setup allows us to assess long-horizon adaptation across diverse clinical scenarios, model scales, and pretraining paradigms.

\textbf{Evaluation setup.}
Unless otherwise specified, all models are trained on 3-minute ECG recordings and evaluated on 3-minute, 2-minute, 1-minute, 30-second, and 10-second inputs without retraining. This protocol reflects realistic deployment settings in which the available monitoring duration may vary~\cite{yang2023biot}, while keeping the focus on practical extension beyond the original 10-second duration. 

\textbf{Backbone models.}
We use four backbone configurations from three representative ECG foundation-model families: CSFM~\cite{gu2025sensing} (Tiny and Base), MERL~\cite{liu2024zero}, and ECG-JEPA~\cite{kim2026jepa}. These backbones vary in model scale and pretraining objective, providing a diverse test bed for evaluating the robustness and generality of the proposed framework. 

\subsection{Datasets \& Tasks}
\label{sec:datasets_tasks}

We consider three datasets covering four downstream task settings in total. All datasets are preprocessed using a standardized pipeline before training and evaluation. 

\textbf{VTaC}~\cite{lehman2023vtac, PhysioNet-vtac-1.0} is a critical care benchmark for reducing false ventricular tachycardia (VT) alarms. We use the final 3 minutes preceding the alarm onset and formulate the task as binary classification of true versus false VT alarms.

\textbf{MC-MED}~\cite{kansal2025mcmed, PhysioNet-mc-med-1.0.1} contains emergency department records and is used for two prediction tasks: \textit{ED disposition} prediction, formulated as binary classification of \textit{Discharge} versus \textit{Admission}, and \textit{triage acuity} prediction, formulated as five-class classification. For both tasks, we use the lead-II ECG signal and extract the final 3 minutes of the selected segment.

\textbf{CPSC2021}~\cite{PhysioNet-cpsc2021-1.0.0} is a long-term Holter benchmark targeting paroxysmal atrial fibrillation (AF) events. We formulate it as a binary AF detection task using the lead-II signal and segment long recordings into non-overlapping 3-minute windows.

\subsection{Baseline Methods}

We compare the proposed method with two groups of baseline strategies: \textit{Snapshot Aggregation} includes Token Pooling and Logit Pooling. These methods partition each long ECG into non-overlapping 10-second windows and aggregate either frozen backbone features or window-level logits. \textit{Fine-tuning-based Adaptation} includes Full Fine-tuning, Linear Probing, Partial Tuning, Bias Tuning, Adapter, Visual Prompt Tuning (VPT)~\cite{chen2022adaptformer,pfeiffer2021adapterfusion,jia2022visual}, and an LSTM-based baseline. These methods adapt either the pretrained model parameters or the resulting sequence of snapshot features for longer recordings. For baselines that require longer inputs, positional embeddings are extended by interpolation.

%% file: section/results.tex
\textbf{Main Results on Long-Horizon ECG Tasks.}
Table~\ref{tab:main_results} summarizes the adaptation results across four backbone configurations and four downstream task settings. For each task, we report both performance at 3-minute and the average across all evaluated inference durations (3-minute, 2-minute, 1-minute, 30-second, and 10-secon). Overall, the proposed method is consistently among the strongest approaches across tasks and backbones, and in most cases outperforms both snapshot aggregation baselines and lightweight adaptation methods.

\input{tables/main_results_summary}
A clear pattern also emerges across methods. Snapshot aggregation strategies such as logit pooling and token pooling provide competitive baselines in several settings, but they are typically outperformed in longer temporal context across tasks and backbones. This suggests that simple aggregation over short-window predictions or features is often insufficient for robust long-horizon adaptation.

The backbone-specific trends are also informative. On the CSFM family and ECG-JEPA, the proposed method achieves the strongest overall performance in most settings, particularly at longer input durations. On MERL, full fine-tuning remains the strongest baseline overall, whereas the proposed method remains highly competitive and often achieves the second-best performance while keeping the pretrained backbone frozen. Taken together, these results indicate that the proposed framework provides a reliable way to extend snapshot ECG foundation models across tasks, backbone scales, and pretraining paradigms.

\textbf{Guidance Source Analysis.} The MERL results motivate a more focused question: how much does semantic extension depend on the quality of the teacher representation? On VTaC, MERL self-guidance is noticeably weaker than full fine-tuning, and snapshot aggregation baselines are also relatively poor. This suggests that the snapshot-level MERL representations may provide a less effective guidance signal for long-horizon extension in this setting.

To examine this effect, we replace MERL self-guidance with external guidance from stronger backbone models while keeping the MERL student unchanged. As shown in Table~\ref{tab:guidance_vtac}, all three external guidance sources consistently improve over MERL self-guidance across inference durations, with the gains becoming more pronounced beyond 30\,s. Figure~\ref{fig:tsne_vtac} further provides qualitative evidence that the different backbones induce different feature-space structures on VTaC, which is consistent with their different effectiveness as guidance sources. Together, these results suggest that long-horizon semantic extension depends, at least in part, on the informativeness of the underlying snapshot-level teacher representation.

\input{tables/merl_cfsm_guided}

\begin{figure*}[t]
    \centering
    \includegraphics[width=\linewidth]{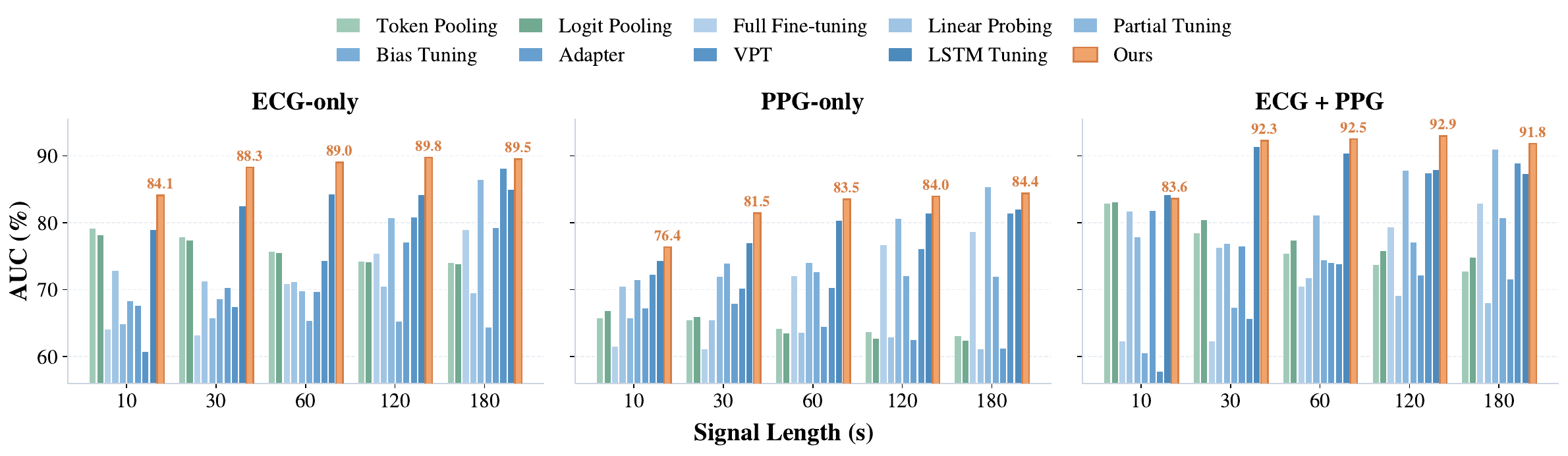}
    \caption{\textbf{Generalization across signal modalities on VTaC.} We evaluate the proposed framework using CSFM-Tiny under ECG-only, PPG-only, and joint ECG+PPG settings. Across all three modality configurations, the proposed method consistently outperforms the baseline adaptation strategies, indicating that the long-horizon extension is not specific to ECG alone.}
    \label{figure:results_csfm_vtac}
    \vspace{-18pt}
\end{figure*}

\textbf{Modality-Agnostic Analysis.}
The long-horizon extension problem is not unique to ECG. Other physiological signals are also commonly modeled using short snapshot windows despite containing clinically meaningful information over longer durations. To test whether the proposed framework generalizes beyond ECG-only inputs, we evaluate CSFM-Tiny on VTaC using ECG-only, PPG-only, and joint ECG+PPG settings. Figure~\ref{figure:results_csfm_vtac} shows that our method consistently outperforms the baseline adaptation strategies across all three modality configurations. This result suggests that the proposed framework is not tied to a single physiological modality, but instead addresses a broader limitation of snapshot-pretrained physiological models.

\textbf{Ablation Study.} We next examine which components of the proposed framework drive its gains. Table~\ref{tab:ablation_csfm_modulesonly} isolates the effects of the positional encoding strategy, locality-aware learning, and dynamics-aware consistency learning on VTaC using CSFM. Replacing repeated or interpolated positional encodings with the proposed global positional encoding already yields a clear improvement, indicating that structural compatibility with longer recordings is an important component of long-horizon extension. Adding either the locality-aware objective or the dynamics-aware objective provides further gains, while combining both produces the strongest overall performance across input durations.

\input{tables/ablation}

\textbf{Affinity-guided dropping.} We further analyze the consistency module through affinity-guided segment dropping. Compared with random dropping, affinity-guided dropping is generally more effective, supporting the role of similarity-based segment structure in the dynamics-aware objective. Taken together, these ablations show that the gains of the proposed framework arise not from simple input resizing or generic tuning alone, but from the combination of structural extension and semantically informed temporal adaptation.



%% file: tables/main_results_summary.tex
\definecolor{TableBlockGray}{RGB}{229,233,238}
\definecolor{TableOursBlue}{RGB}{219,238,252}
\newcommand{\std}[1]{{\scriptsize\textcolor{black!60}{ (#1) }}}

\begin{table*}[t]
\centering
\caption{
\textbf{Main adaptation results across ECG foundation models and downstream tasks.}
For each task, we report performance at the longest evaluation horizon (180s) together with the average across five input lengths in the format \textbf{Avg. (Std.)}. 
\textbf{Best} results are shown in bold and \underline{second-best} results are underlined.
}
\label{tab:main_results}
\vspace{-2mm}
\scriptsize
\setlength{\tabcolsep}{3.2pt}
\renewcommand{\arraystretch}{1.08}
\resizebox{0.88\textwidth}{!}{%
\begin{tabular}{l cc cc cc cc c}
\toprule
\multirow{2}{*}{\textbf{Method}} &
\multicolumn{2}{c}{\textbf{VTaC}} &
\multicolumn{2}{c}{\textbf{MC-MED (ED)}} &
\multicolumn{2}{c}{\textbf{MC-MED (Acuity)}} &
\multicolumn{2}{c}{\textbf{CPSC2021}} &
\multirow{2}{*}{\textbf{Overall Avg.}} \\
\cmidrule(lr){2-3}
\cmidrule(lr){4-5}
\cmidrule(lr){6-7}
\cmidrule(lr){8-9}
& \textbf{180s} & \textbf{Avg. \std{Std.}}
& \textbf{180s} & \textbf{Avg. \std{Std.}}
& \textbf{180s} & \textbf{Avg. \std{Std.}}
& \textbf{180s} & \textbf{Avg. \std{Std.}}
& \\
\midrule

\rowcolor{TableBlockGray}
\multicolumn{10}{c}{\textbf{\textit{CSFM-Tiny}}} \\

Token Pooling
& 74.05 & 76.24 \std{2.28}
& 70.80 & 68.72 \std{2.69}
& 61.00 & 60.00 \std{1.03}
& 96.81 & 96.59 \std{0.39}
& 75.39 \\

Logit Pooling
& 73.83 & 75.83 \std{1.96}
& 71.15 & \underline{69.28} \std{2.26}
& 60.13 & 59.96 \std{0.48}
& 97.23 & 97.26 \std{0.16}
& 75.58 \\

Full Fine-tuning
& 79.01 & 70.54 \std{6.89}
& 70.81 & 68.09 \std{3.05}
& \underline{64.80} & \underline{63.07} \std{1.18}
& 94.24 & 93.23 \std{1.91}
& 73.73 \\

Linear Probing
& 69.50 & 71.10 \std{1.25}
& 69.38 & 66.89 \std{2.67}
& 60.66 & 60.08 \std{0.99}
& 96.46 & 96.91 \std{0.57}
& 73.75 \\

Partial Tuning
& 86.49 & 73.57 \std{9.59}
& 70.91 & 68.47 \std{3.79}
& 63.43 & 63.06 \std{1.50}
& 95.48 & 94.87 \std{1.57}
& 74.99 \\

Bias Tuning
& 64.43 & 66.43 \std{1.94}
& 68.13 & 66.28 \std{1.93}
& 56.71 & 54.49 \std{2.42}
& 98.46 & 97.62 \std{1.33}
& 71.21 \\

Adapter
& 79.33 & 72.83 \std{5.08}
& 71.24 & 68.60 \std{3.01}
& 61.55 & 60.53 \std{0.94}
& 98.01 & 96.24 \std{2.99}
& 74.55 \\

VPT
& \underline{88.18} & 74.33 \std{10.78}
& 71.19 & 67.29 \std{5.16}
& 64.50 & 61.22 \std{3.17}
& 98.67 & 96.37 \std{3.94}
& 74.80 \\

LSTM Tuning
& 84.97 & \underline{83.01} \std{2.39}
& \underline{71.31} & 65.18 \std{7.55}
& 61.97 & 58.80 \std{4.55}
& \underline{98.75} & \underline{98.09} \std{0.63}
& \underline{76.27} \\

\rowcolor{TableOursBlue}
\textbf{Ours}
& \textbf{89.50} & \textbf{88.13} \std{2.32}
& \textbf{72.02} & \textbf{69.79} \std{3.18}
& \textbf{65.18} & \textbf{63.35} \std{2.04}
& \textbf{98.86} & \textbf{98.63} \std{0.79}
& \textbf{79.98} \\

\midrule

\rowcolor{TableBlockGray}
\multicolumn{10}{c}{\textbf{\textit{CSFM-Base}}} \\

Token Pooling
& 77.94 & 79.84 \std{2.09}
& 71.23 & 69.30 \std{2.51}
& 66.82 & 64.76 \std{2.36}
& 97.02 & 97.33 \std{0.40}
& 77.81 \\

Logit Pooling
& 77.83 & 79.64 \std{1.98}
& 71.59 & \underline{69.90} \std{2.46}
& 67.07 & 64.85 \std{2.46}
& 97.54 & 97.52 \std{0.23}
& 77.98 \\

Full Fine-tuning
& 84.23 & 72.46 \std{9.58}
& 72.30 & 69.83 \std{3.13}
& 65.09 & 63.47 \std{1.46}
& 95.66 & 95.66 \std{1.13}
& 75.35 \\

Linear Probing
& 74.35 & 77.50 \std{2.81}
& 70.44 & 67.86 \std{2.64}
& 66.53 & 64.26 \std{2.32}
& 96.89 & 96.86 \std{0.25}
& 76.62 \\

Partial Tuning
& 88.81 & 79.69 \std{8.23}
& 71.47 & 69.29 \std{3.41}
& 67.25 & 65.06 \std{2.11}
& 96.21 & 95.54 \std{1.00}
& 77.40 \\

Bias Tuning
& 70.35 & 73.35 \std{2.02}
& 71.36 & 68.87 \std{3.16}
& 60.66 & 58.90 \std{1.77}
& 98.39 & 97.23 \std{1.39}
& 74.59 \\

Adapter
& 88.43 & 78.95 \std{7.96}
& 71.90 & 69.22 \std{3.45}
& 65.91 & 63.72 \std{2.00}
& 98.41 & 96.85 \std{2.87}
& 77.19 \\

VPT
& \underline{90.24} & 80.72 \std{7.93}
& \underline{72.39} & 68.76 \std{5.25}
& 66.00 & 63.78 \std{3.03}
& 98.57 & 96.57 \std{2.89}
& 77.46 \\

LSTM Tuning
& 86.93 & \underline{85.22} \std{1.33}
& 72.03 & 67.77 \std{6.33}
& \underline{68.14} & \underline{65.18} \std{2.95}
& \underline{98.74} & \underline{98.31} \std{0.47}
& \underline{79.12} \\

\rowcolor{TableOursBlue}
\textbf{Ours}
& \textbf{91.23} & \textbf{89.62} \std{1.53}
& \textbf{72.89} & \textbf{70.36} \std{3.43}
& \textbf{68.89} & \textbf{66.00} \std{2.65}
& \textbf{99.56} & \textbf{98.46} \std{1.08}
& \textbf{81.11} \\

\midrule

\rowcolor{TableBlockGray}
\multicolumn{10}{c}{\textbf{\textit{ECG-JEPA}}} \\

Token Pooling
& 76.76 & 78.54 \std{1.56}
& 69.75 & 67.37 \std{2.74}
& 61.38 & 60.33 \std{1.30}
& 90.75 & 90.68 \std{0.55}
& 74.23 \\

Logit Pooling
& 76.23 & 78.10 \std{1.65}
& 69.23 & 67.67 \std{2.46}
& 61.23 & 60.11 \std{1.28}
& 90.14 & 90.47 \std{0.57}
& 74.09 \\

Full Fine-tuning
& 88.32 & 82.96 \std{5.44}
& 67.96 & 65.44 \std{2.27}
& 62.83 & 60.65 \std{1.56}
& 88.67 & 89.59 \std{0.83}
& 74.66 \\

Linear Probing
& 77.21 & 79.00 \std{1.45}
& 70.32 & 66.49 \std{3.04}
& 63.71 & 61.90 \std{1.46}
& 88.83 & 89.33 \std{0.34}
& 74.18 \\

Partial Tuning
& 88.89 & \underline{84.32} \std{5.27}
& 70.19 & 67.79 \std{2.92}
& 64.59 & 62.54 \std{2.20}
& 88.33 & 89.05 \std{0.76}
& 75.93 \\

Bias Tuning
& 86.83 & 83.68 \std{5.53}
& \underline{72.14} & \underline{69.53} \std{2.77}
& 65.50 & \underline{64.06} \std{1.62}
& 85.41 & 84.86 \std{2.20}
& 75.53 \\

Adapter
& 70.57 & 67.91 \std{2.44}
& 63.06 & 60.68 \std{2.34}
& 60.77 & 60.17 \std{0.60}
& 75.88 & 73.17 \std{4.10}
& 65.48 \\

VPT
& \underline{89.87} & 82.44 \std{7.97}
& \textbf{72.48} & 67.29 \std{6.11}
& \underline{66.18} & 63.89 \std{3.53}
& \underline{93.65} & 84.85 \std{10.15}
& 74.62 \\

LSTM Tuning
& 85.17 & 82.66 \std{2.67}
& 70.72 & 68.39 \std{3.15}
& 65.18 & 62.93 \std{2.35}
& 91.26 & \underline{90.90} \std{0.49}
& \underline{76.22} \\

\rowcolor{TableOursBlue}
\textbf{Ours}
& \textbf{91.02} & \textbf{85.65} \std{4.91}
& \textbf{72.48} & \textbf{70.69} \std{3.35}
& \textbf{67.53} & \textbf{64.89} \std{2.63}
& \textbf{94.66} & \textbf{90.92} \std{4.25}
& \textbf{78.04} \\

\midrule

\rowcolor{TableBlockGray}
\multicolumn{10}{c}{\textbf{\textit{MERL}}} \\

Token Pooling
& 71.56 & 69.47 \std{3.97}
& 68.79 & 65.49 \std{3.85}
& 66.82 & 64.76 \std{2.36}
& 88.31 & 86.59 \std{3.38}
& 71.58 \\

Logit Pooling
& 71.37 & 69.32 \std{3.88}
& 68.81 & 65.47 \std{3.78}
& 67.07 & 64.85 \std{2.46}
& 86.56 & 85.79 \std{2.59}
& 71.36 \\

Full Fine-tuning
& \textbf{88.31} & \textbf{82.08} \std{7.39}
& 67.30 & 65.03 \std{2.81}
& 65.09 & 63.47 \std{1.46}
& \textbf{95.72} & \textbf{91.50} \std{7.41}
& \textbf{75.52} \\

Linear Probing
& 74.91 & 71.94 \std{3.98}
& 68.07 & 64.64 \std{5.25}
& 66.53 & 64.26 \std{2.32}
& 84.52 & 82.86 \std{2.40}
& 70.93 \\

Partial Tuning
& 79.06 & 71.59 \std{6.79}
& 68.41 & 64.98 \std{4.60}
& 67.25 & 65.06 \std{2.11}
& 88.71 & 87.32 \std{2.78}
& 72.24 \\

Bias Tuning
& 80.63 & 75.62 \std{6.72}
& 69.12 & \underline{65.57} \std{4.52}
& 60.66 & 58.90 \std{1.77}
& \underline{94.10} & 88.26 \std{6.51}
& 72.09 \\

Adapter
& 74.72 & 71.16 \std{4.38}
& 68.36 & 65.39 \std{4.77}
& 65.91 & 63.72 \std{2.00}
& 89.39 & 88.52 \std{2.24}
& 72.20 \\

VPT
& 80.74 & 71.94 \std{8.67}
& \underline{69.74} & 64.29 \std{5.69}
& 66.00 & 63.78 \std{3.03}
& 91.75 & 85.59 \std{8.69}
& 71.40 \\

LSTM Tuning
& \underline{83.05} & \underline{78.55} \std{5.44}
& 67.74 & 63.86 \std{3.97}
& \underline{68.14} & \underline{65.18} \std{2.95}
& 93.55 & 89.01 \std{5.96}
& 74.15 \\

\rowcolor{TableOursBlue}
\textbf{Ours}
& 82.60 & 76.76 \std{9.16}
& \textbf{69.92} & \textbf{65.76} \std{5.32}
& \textbf{68.89} & \textbf{66.00} \std{2.65}
& 93.49 & \underline{89.73} \std{5.89}
& \underline{74.56} \\

\bottomrule
\end{tabular}%
}
\vspace{-8mm}
\end{table*}

%% file: tables/merl_cfsm_guided.tex
\begin{figure*}[t]
\centering

\begin{minipage}[t]{0.45\textwidth}
    \raggedright
    \vspace{0pt}
    \scriptsize
    \setlength{\tabcolsep}{4pt}
    \renewcommand{\arraystretch}{1.12}
    \captionof{table}{\textbf{Guidance source comparison for MERL on VTaC.}
    Results are reported as AUC (\%) across different inference durations.}
    \label{tab:guidance_vtac}

    \vspace{1mm}
    \resizebox{\linewidth}{!}{%
    \begin{tabular}{lccccc}
    \toprule
    \textbf{Guidance} & \textbf{10s} & \textbf{30s} & \textbf{60s} & \textbf{120s} & \textbf{180s} \\
    \midrule
    MERL Self-guided & 61.23 & 76.03 & 80.27 & 83.68 & 82.60 \\
    CSFM-Tiny guided & 62.67 & 79.64 & 83.46 & 84.64 & 85.19 \\
    CSFM-Base guided & \textbf{63.16} & \textbf{81.33} & \textbf{84.51} & 84.99 & \textbf{86.07} \\
    ECG-JEPA guided  & 63.07 & 80.61 & 84.17 & \textbf{85.03} & 85.78 \\
    \bottomrule
    \end{tabular}%
    }
\end{minipage}
\hspace{0.01\textwidth}
\begin{minipage}[t]{0.48\textwidth}
    \raggedleft
    \vspace{0pt}

    \begin{minipage}[t]{0.245\linewidth}
        \centering
        \includegraphics[width=\linewidth]{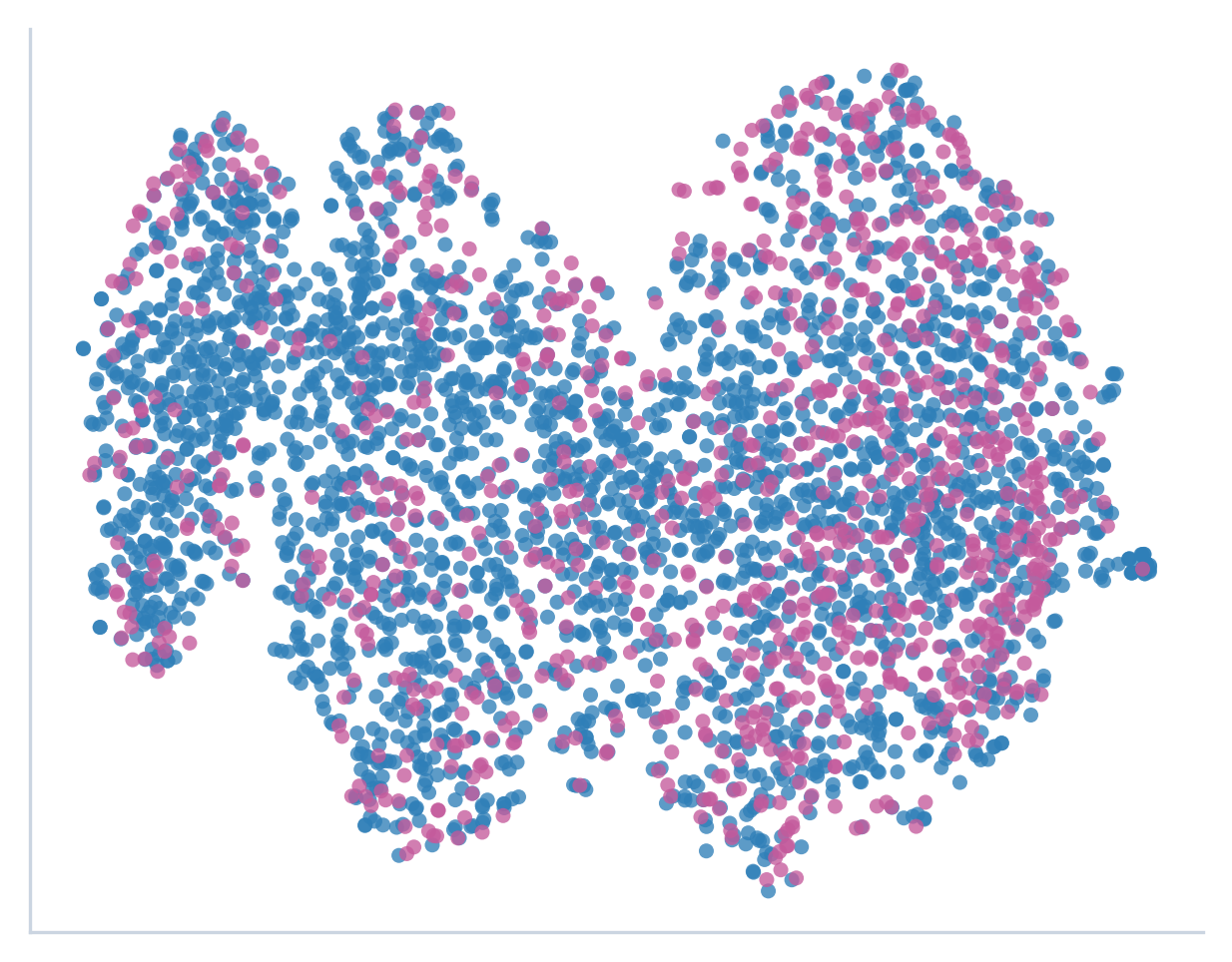}
        {\scriptsize MERL}
    \end{minipage}%
    \begin{minipage}[t]{0.245\linewidth}
        \centering
        \includegraphics[width=\linewidth]{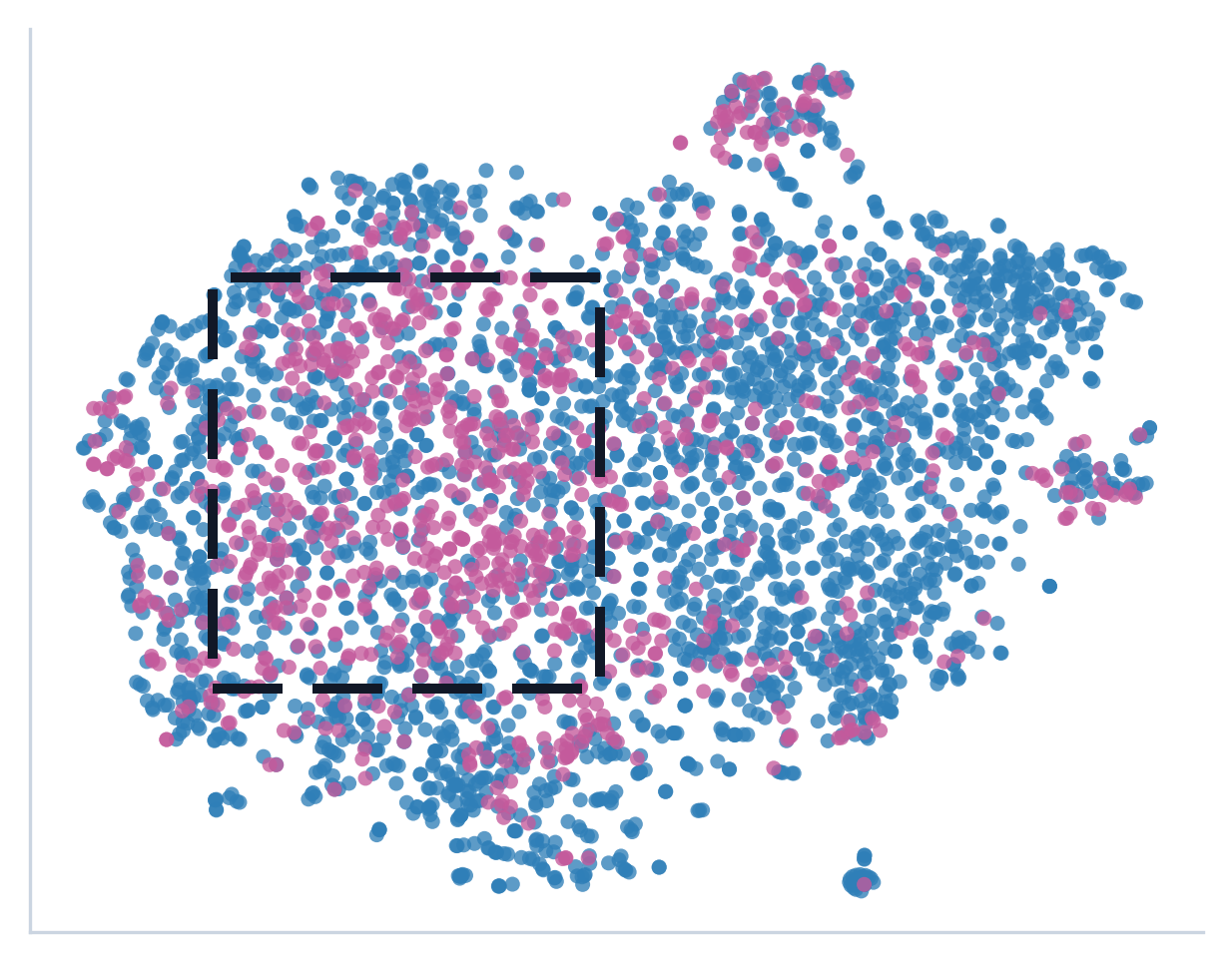}
        {\scriptsize CSFM-Tiny}
    \end{minipage}%
    \begin{minipage}[t]{0.245\linewidth}
        \centering
        \includegraphics[width=\linewidth]{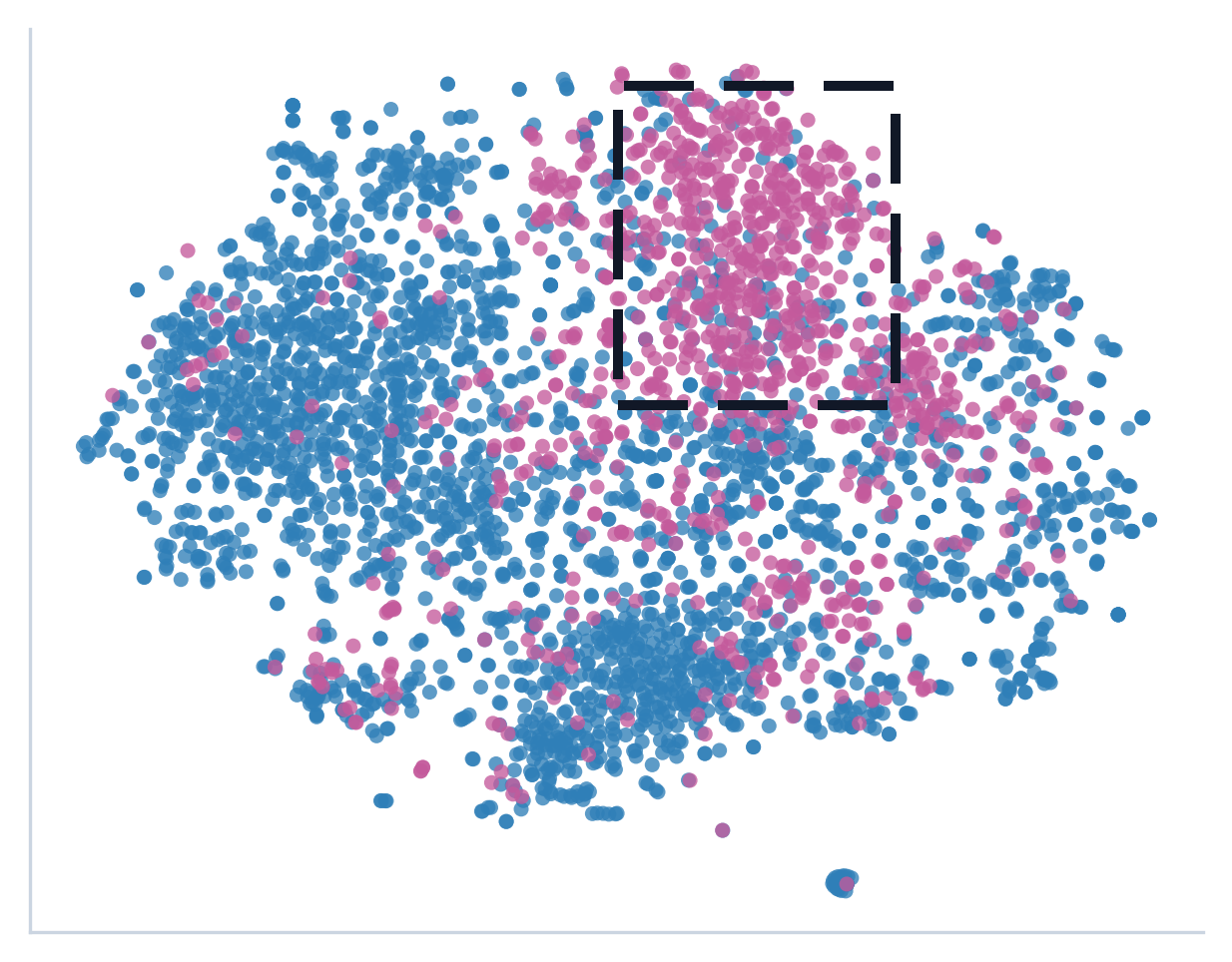}
        {\scriptsize CSFM-Base}
    \end{minipage}%
    \begin{minipage}[t]{0.245\linewidth}
        \centering
        \includegraphics[width=\linewidth]{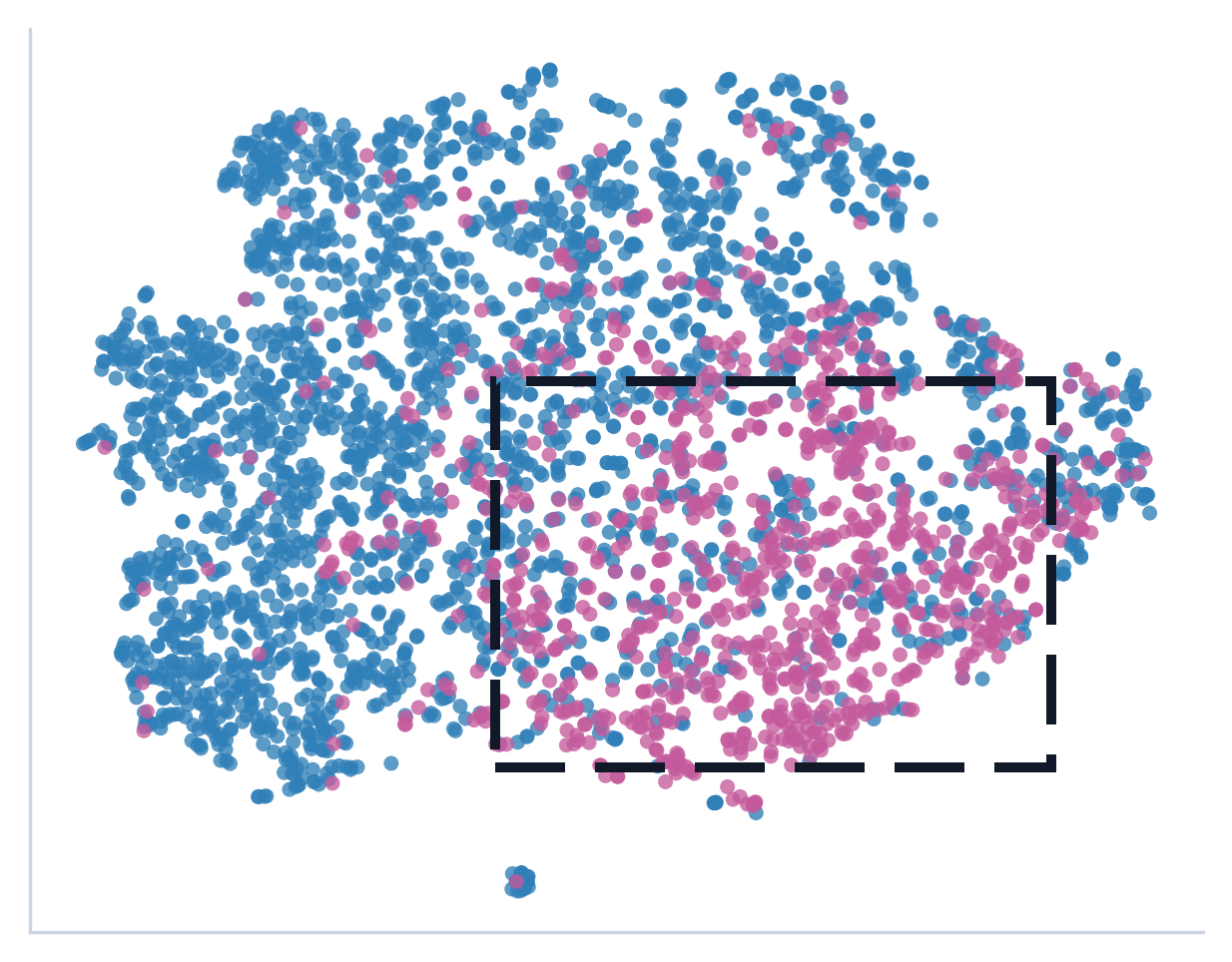}
        {\scriptsize ECG-JEPA}
    \end{minipage}

    \vspace{-1.6mm}
    \captionof{figure}{\textbf{t-SNE visualizations of feature space on VTaC.}
    \protect\textcolor{tsneblue}{Blue} and \protect\textcolor{tsnepurple}{purple} denote different classes.
    The dashed boxes highlight regions with clearer class separation, indicating stronger teacher model.}
    \label{fig:tsne_vtac}
\end{minipage}

\vspace{-2mm}
\end{figure*}

%% file: tables/ablation.tex
\begin{wraptable}[19]{r}{0.5\columnwidth}
\vspace{-5mm}
\begin{minipage}{\linewidth}
\centering
\captionsetup{type=table}
\captionof{table}{\textbf{Ablation of adaptation components} for CSFM on VTaC (AUC \%).}
\label{tab:ablation_csfm_modulesonly}
\small
\setlength{\tabcolsep}{5pt}
\renewcommand{\arraystretch}{1.12}
\resizebox{\linewidth}{!}{%
\begin{tabular}{ccc|ccccc}
\toprule
\textbf{PE Strategy} & $\mathcal{L}_{\text{local}}$ & $\mathcal{L}_{\text{dyn}}$ & 10s & 30s & 60s & 120s & 180s \\
\midrule
Repeat & \xmark & \xmark & 70.43 & 73.09 & 70.60 & 72.43 & 71.45 \\
Interpolate & \xmark & \xmark & 60.78 & 67.44 & 74.37 & 80.86 & 88.18 \\
Global & \xmark & \xmark & 78.13 & 83.96 & 87.45 & 89.08 & 88.92 \\
Global & \cmark & \xmark & 83.62 & 87.13 & 88.42 & 89.30 & 89.00 \\
Global & \xmark & \cmark & 80.01 & 85.90 & 87.80 & 88.86 & 88.90 \\
Global & \cmark & \cmark & \textbf{84.10} & \textbf{88.29} & \textbf{89.01} & \textbf{89.76} & \textbf{89.50} \\
\bottomrule
\end{tabular}}
\end{minipage}

\vspace{3mm}

\begin{minipage}{\linewidth}
\centering
\captionsetup{type=table}
\captionof{table}{\textbf{Affinity-guided dropping} compared with random dropping (AUC \%).}
\label{tab:affinity_drop}
\scriptsize
\setlength{\tabcolsep}{3.5pt}
\renewcommand{\arraystretch}{1.1}
\resizebox{\linewidth}{!}{%

\begin{tabular}{llccccc}
\toprule
\textbf{Model} & \textbf{Strategy} & \textbf{10s} & \textbf{30s} & \textbf{60s} & \textbf{120s} & \textbf{180s} \\
\midrule
\multirow{2}{*}{CSFM-Tiny}
& Affinity Drop & \textbf{84.10} & \textbf{88.29} & \textbf{89.01} & 89.76 & \textbf{89.50} \\
& Random Drop   & 84.00 & 87.19 & 88.35 & \textbf{90.03} & 89.01 \\

\multirow{2}{*}{CSFM-Base}
& Affinity Drop & \textbf{87.38} & \textbf{89.15} & \textbf{89.50} & \textbf{90.85} & \textbf{91.23} \\
& Random Drop   & 86.73 & 88.01 & 88.93 & 90.16 & 90.83 \\

\multirow{2}{*}{ECG-JEPA}
& Affinity Drop & \textbf{78.86} & \textbf{82.93} & \textbf{86.09} & \textbf{89.36} & \textbf{91.02} \\
& Random Drop   & 75.92 & 80.79 & 84.87 & 88.92 & 90.01 \\

\multirow{2}{*}{MERL}
& Affinity Drop & \textbf{61.23} & \textbf{76.03} & \textbf{80.27} & \textbf{83.68} & \textbf{82.60} \\
& Random Drop   & 58.20 & 72.54 & 78.67 & 80.23 & 81.20 \\
\bottomrule
\end{tabular}}

\end{minipage}
\vspace{-20mm}
\end{wraptable}

%% file: section/conclusion.tex

In this work, we studied how to extend snapshot ECG foundation models beyond their original fixed-length input setting. Without full retraining, we introduced a lightweight plug-in adaptation module that enables one-time extension to longer and variable-length recordings. Across multiple tasks, datasets, and backbone families, the proposed method achieves consistently strong performance, improving over snapshot aggregation baselines and remaining competitive across diverse adaptation strategies.

A key feature of the proposed method is that the extension module is trained at the longest available horizon and then evaluated across a range of different input lengths. The resulting performance indicates that the adapted models can generalize effectively across variable-duration inputs. However, extending snapshot foundation models to arbitrary or substantially longer horizons under practical memory and computational constraints remains an open challenge. We view more inference-efficient mechanisms for longer-horizon adaptation as an important direction for future work.